\pdfoutput=1

\documentclass[11pt]{article}
\usepackage{graphicx}
\usepackage{subfigure}
\usepackage{ACL2023}
\usepackage{amsfonts,amssymb}
\usepackage{times}
\usepackage{latexsym}

\usepackage[T1]{fontenc}

\usepackage[utf8]{inputenc}

\usepackage{microtype}

\usepackage{inconsolata}
\usepackage{tablefootnote}
\usepackage{lipsum}

\title{Analyzing and Reducing the Performance Gap in Cross-Lingual Transfer with Fine-tuning Slow and Fast}

\author{
Yiduo Guo$^{1,}$\footnotemark[1],~~Yaobo Liang$^{2}$,~~Dongyan Zhao$^{1}$\footnotemark[2],~~Bing Liu$^{3}$,~~Duan Nan$^{2}$\\ 
$^1$Wangxuan Institute of Computer Technology, Peking University\\
$^2$Microsoft Research Asia\\
$^3$Department of Computer Science, University of Illinois at Chicago\\
\texttt{yiduo@stu.pku.edu.cn,yaobo.liang@microsoft.com,  zhaody@pku.edu.cn}\\\texttt{nanduan@microsoft.com,liub@uic.edu}\\
}

\begin{document}
\maketitle
\renewcommand{\thefootnote}{\fnsymbol{footnote}}
\footnotetext[1]{Work done during internship at Microsoft Research Asia}
\footnotetext[2]{Corresponding author}
\begin{abstract}
Existing research has shown that a multilingual pre-trained language model fine-tuned with one (source) language also performs well on downstream tasks for non-source languages, even though no fine-tuning is done on these languages. However, there is a clear gap between the performance of the source language and that of the non-source languages. This paper analyzes the fine-tuning process, discovers when the performance gap changes and identifies which network weights affect the overall performance most. Additionally, the paper seeks to answer to what extent the gap can be reduced by reducing forgetting. Based on the analysis results, a method named \textbf{Fine-tuning slow and fast} with four training policies is proposed to address these issues. Experimental results show the proposed method outperforms baselines by a clear margin.
\end{abstract}

\section{Introduction}
Multilingual pre-trained language models (LMs), such as mBERT~\cite{devlin2018bert} and XLM-R~\cite{conneau2019unsupervised} have shown strong Zero-Shot Cross-Lingual transfer capabilities. Such a model $F$ is usually pre-trained with unlabeled corpora $D$ in multiple languages $S$ to enable the model to learn cross-lingual knowledge $H_{cross}^{pre}$. To adapt to a downstream task, the pre-trained LM $F$ is typically fine-tuned with a supervised dataset $D_{\hat{s}}$ of the downstream task $T$ in one source language $\hat{s}\in S$ due to data scarcity in other languages. When the fine-tuned model $\overline{F}$ is applied to the test set of the same task in the source language $\hat{s}$, it achieves strong performance $P_{\hat{s}}$. Interestingly, when $\overline{F}$ is applied to non-source languages, it can also achieve good performance ~\cite{conneau2019unsupervised}. We denote the average performance on the test sets of other languages than $\hat{s}$ as $P_{S/\hat{s}}$. 
\begin{figure}[h]
\centering 
\vspace{-2mm}

\includegraphics[height=0.35\textwidth,width=0.45\textwidth]{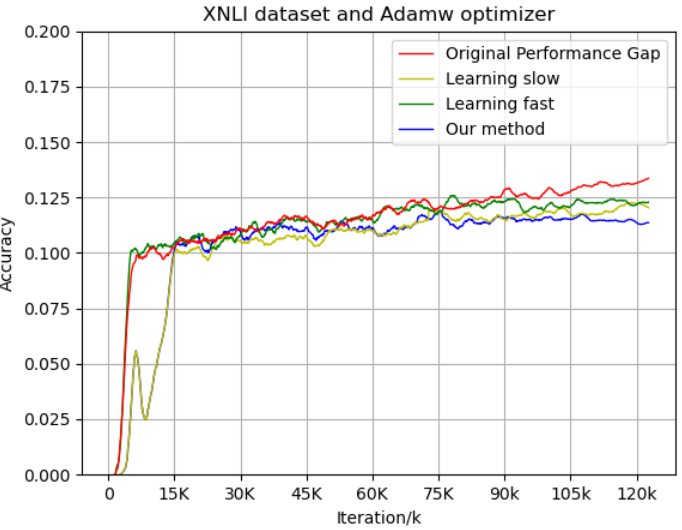} 
\vspace{-3mm}
\caption{The performance gap $P_{\hat{s}}-P_{S/\hat{s}}$  every hundred updates on the XNLI dataset. 
'Original performance gap' means that we directly fine-tune the model, and 'Fine-tuning slow'/'fine-tuning fast'/'our method' means that we use the fine-tuning slow algorithm/fine-tuning fast algorithm/the combination of both algorithms respectively to fine-tune the model.}
\label{Fig.gap_first}
\vspace{-4mm}
\end{figure}
However, the gap $P_{\hat{s}}-P_{S/\hat{s}}$ is quite large (e.g., 13 percent for the XNLI dataset in Figure 1). {\color{black}One potential reason is that: during the fine-tuning of the model, the performance of non-source languages firstly increases with the performance of source language, then the arising of the performance of non-source languages becomes slower than that of the performance of source language as the forgetting of cross-lingual knowledge, resulting in a larger gap.} 
Inspired by the study of \textit{catastrophic forgetting} (CF) phenomenon in continual learning (CL), we introduce a classical concept in CL here to help solve our problem: the dilemma of \textit{plasticity vs. stability}. 


\textbf{Plasticity vs Stability.} In CL \cite{kirkpatrick2017overcoming}, the learner needs to learn a sequence of different tasks incrementally. 
Plasticity means learning and performing well on the new task and stability means maintaining the learned knowledge of the previous tasks.
 The learner needs to find a balance between plasticity and stability because too much plasticity (e.g, changing the entire model drastically) causes serious CF of the learned knowledge, and too much stability (e.g. freezing the whole model) makes the model can not learn new things. Fine-tuning a multi-lingual LM $F$ using only the corpus of one source language also meets this balance dilemma. Thus, Fine-tuning LM $F$ needs to protect the cross-lingual knowledge $H_{cross}^{pre}$ (stability) and also learn the new task knowledge $H_{task}^{new}$ via fine-tuning to adapt to the specific downstream task (plasticity). However, further analysis of the performance gap and the dilemma of plasticity and stability in cross-lingual fine-tuning is needed.

\begin{figure*}[h!]
\centering 
\vspace{-2mm}

\includegraphics[height=0.25\textwidth,width=1\textwidth]{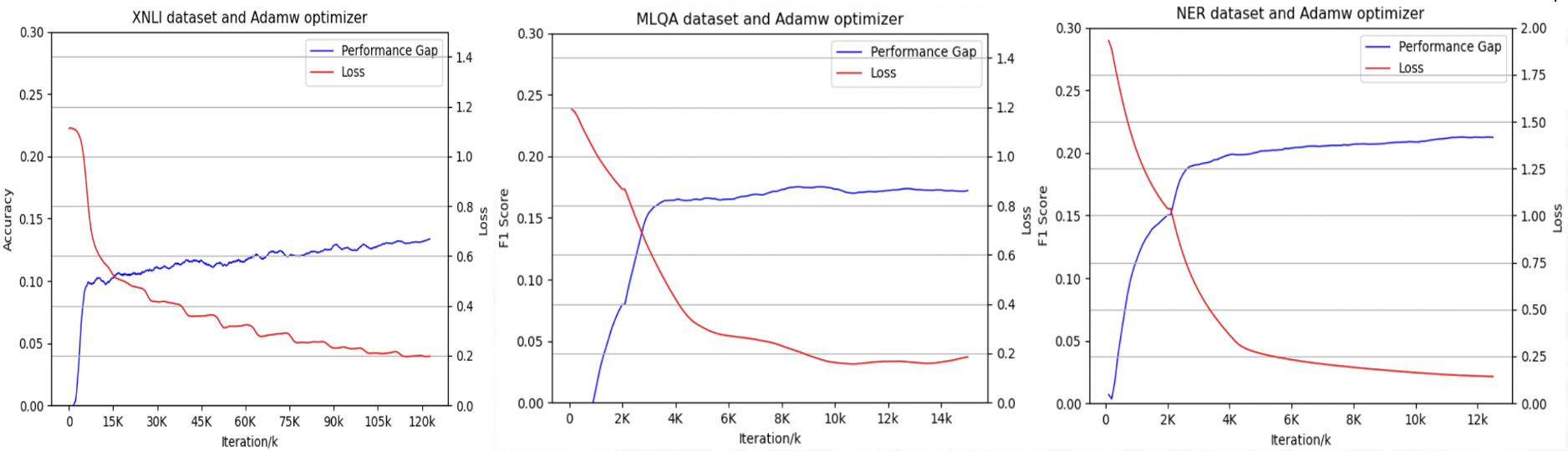} 
\vspace{-5mm}
\caption{We record the loss and the performance gap between English and non-source languages every hundred updates over three different datasets and plot the curves in this figure.}
\label{Fig.when}
\vspace{-4mm}
\end{figure*}

This paper further investigates three research questions: 1) When does the performance gap arise during fine-tuning using a labeled source language corpus? 
2) Where is the most important part of the pre-trained model for achieving strong zero-shot cross-lingual performances? 3) To what extent can we reduce the performance gap by reducing the forgetting of $H_{cross}^{pre}$? 
Based on the experiments on three datasets of different downstream tasks, our analysis found that the performance gap arises significantly in the initial fine-tuning phase and increases slowly in the later phase (see Figure~\ref{Fig.when}).
Feed-forward weights in the bottom four layers are the key weights for the cross-lingual knowledge (See Figure~\ref{Fig.where} and Table~\ref{tab:gapattfeed}) and should be updated slowly to avoid forgetting $H^{pre}_{cross}$. Attention weights in the top two layers have the pre-training task (e.g., Masked-Language Modeling) knowledge $H^{pre}_{task}$ and $H^{pre}_{task}$ is useless for the downstream task. So these weights should be updated fast to encourage forgetting $H^{pre}_{task}$.
We also find that protecting the cross-lingual knowledge by freezing the weights related to it can reduce the performance gap (enough stability) but cannot eliminate the gap completely (See Figure~\ref{Fig.what} ). That means only reducing the forgetting of $H^{pre}_{cross}$ is not enough for solving the performance gap.

\textbf{Un-forgetting vs forgetting.}
Based on the above analysis, we propose a method called \textbf{Fine-tuning slow and Fast algorithm} to mitigate the forgetting of cross-lingual knowledge (stability) and also to selectively forget  the knowledge related to the pre-training task (plasticity) to adapt $F$ to the downstream task in fine-tuning $F$. Note that traditional techniques for solving the forgetting problem in continual learning are not applicable to our setting directly (see the reasons in Sec~\ref{sec.related}). 

The proposed method consists of four learning rate policies. Policies I and II (stability policies) are respectively designed to avoid forgetting of $H_{cross}^{pre}$ in the first fine-tuning stage and to avoid the forgetting of $H_{cross}^{pre}$ based on the tendency of the learning curve in the second fine-tuning stage. Policies III and IV (plasticity policies) are respectively designed to selectively forget the pre-training task knowledge in $H_{task}^{pre}$ in the initial fine-tuning stage where the loss drops drastically and to further encourage forgetting of the pre-training task knowledge $H_{task}^{pre}$ and the learning of $H_{task}^{new}$ in the second fine-tuning stage. 

{\color{blue}
{\color{black}This paper's main contributions are as follows:

(1) We analyze the performance gap in cross-lingual fine-tuning and answer to what extent we can reduce the performance gap by avoiding forgetting cross-lingual knowledge.

(2) We propose a method consisting of four learning rate policies to reduce forgetting of cross-lingual knowledge (stability) and to encourage forgetting of pre-training task-related knowledge (plasticity).

(3) We test our method in multiple datasets under zero and few-shot settings. Compared to the baseline, our method reduces the performance gap (Figure~\ref{Fig.gap_first}(XNLI) and Figure~\ref{Fig.gap_2} in Appendix A (MLQA and NER)) and achieves better overall performance (Table~\ref{tab:accresults}) by protecting the cross-lingual knowledge and learning better task representation. 
}}

\section{Analysis of Performance Gap and Forgetting of Cross-Lingual Knowledge in Fine-Tuning}
This section studies three research questions, i.e., {\color{black}when $P_{\hat{s}}-P_{S/\hat{s}}$ happens and where the weights influence the overall performance mostly?} It also answers to what extent we can reduce the performance gap $P_{\hat{s}}-P_{S/\hat{s}}$ by reducing the forgetting of cross-language knowledge in  $H_{cross}^{pre}$.

\subsection{Overall Setup}
We directly use the multilingual pre-trained model XLM-R~\cite{conneau2019unsupervised} as the base LM due to its strong zero-shot cross-lingual transfer performance. We consider the Cross-lingual Natural Language Inference (XNLI) dataset \cite{conneau2018xnli} which is a cross-lingual textual entailment dataset (classification task). 
Multilingual Question Answering (MLQA) dataset \cite{lewis2019mlqa} which is a multilingual machine reading comprehension task and
NER dataset (named entity recognition task) in
XTREME benchmark \cite{hu2020xtreme}. The metric for MLQA and NER is the F1 score and the metric for XNLI is accuracy. \textbf{All results are the average of 4 random seeds}. We use the zero-shot cross-lingual transfer setting with English as the source language for all experiments. More training details are in Appendix B.

\begin{figure*}[h]
\centering 
\includegraphics[height=0.3\textwidth,width=1\textwidth]{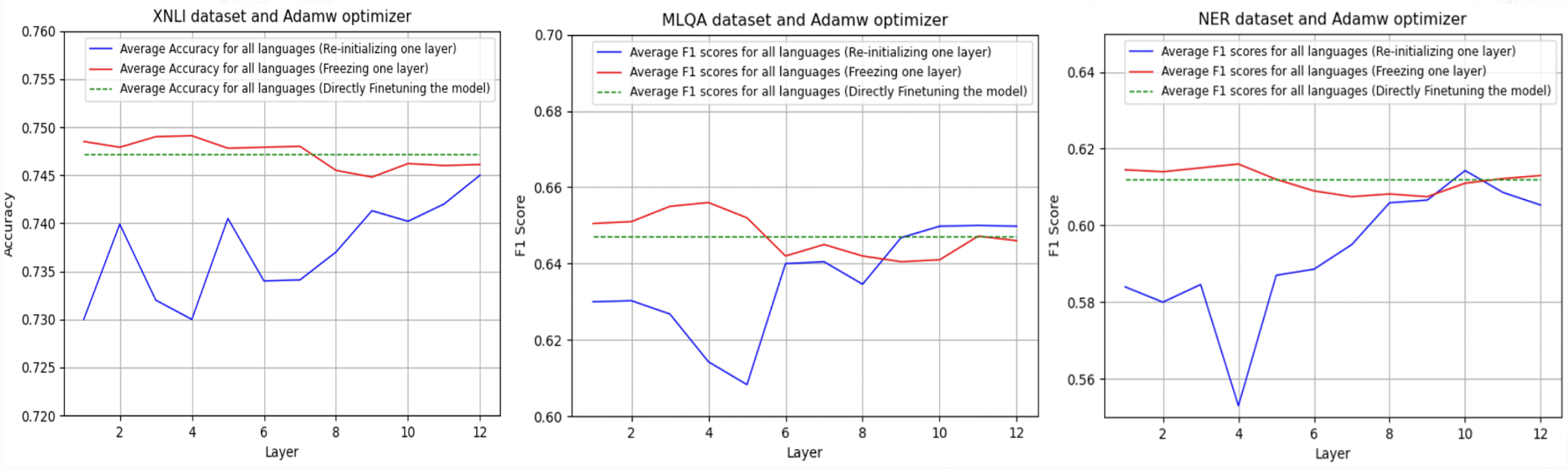} 
\vspace{-4mm}
\caption{For all twelve layers in the pre-trained XLM-R model, we choose one layer and re-initialize its weight before training or freeze its weight during training. We record the final average performance and plot the curve. Y-axis is the metric and X-axis is the index of the layer we chose. The dotted line is the performance of directly fine-tuning model $F$.}
\label{Fig.where}
\end{figure*}
\subsection{When does the Performance Gap Arise?}
We record the loss and calculate the performance gap on the validation set every hundred updates. Figure \ref{Fig.when} shows that the occurrence of the performance gap can be divided into \textbf{two phases: (1) In the first phase $P_1$ (the first 20\% of iterations), the performance gap occurs early and increases dramatically as the loss drops quickly in the initial training stage. (2) In the second phase $P_2$ 
 (the last 80\% iterations), the gap increases but is obviously slower than in the first phase and the loss drops slowly.} 

\subsection{Where is the Knowledge that Helps Cross-Lingual Transfer?}

We use freezing and re-initializing functions to investigate the influence of the weights of each layer on overall performance. Note that we only choose one layer to do re-initializing/freezing operations in each experiment.
Figure \ref{Fig.where} shows that \textbf{the cross-lingual knowledge $H_{cross}^{pre}$ widely exists in the first 10 layers} as re-initializing the weights in any of the ten layers causes performance drop \textbf{and is mainly located in the first four layers} as re-initializing the weights in one of the first four layers makes the performance drops obviously and freezing them boosts the performance. Also interestingly, the pre-trained knowledge in the last two layers has little influence on performance. Sometimes re-initializing one layer in the two layers even makes the performance better than the baseline performance (e.g., for the MLQA dataset). That is because the task of pre-training (e.g. Masked Language Model task) is different from the downstream task and mainly located in the last two layers. We call this kind of knowledge learned from the pre-training task the \textbf{pre-training task} knowledge $H_{task}^{pre}$, which can have a negative transfer to the downstream task.

\subsection{How much can We Reduce the Performance Gap by Reducing Forgetting?}
\begin{figure*}[h]
\centering 

\includegraphics[height=0.3\textwidth,width=1\textwidth]{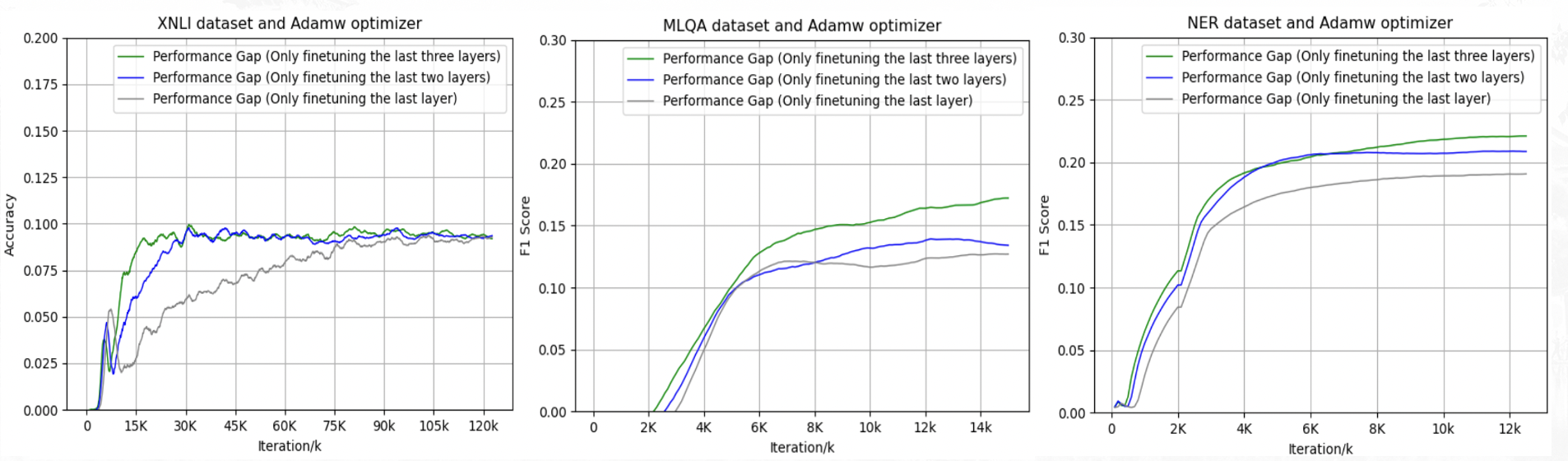} 
\vspace{-5mm}
\caption{We fine-tune only the last one/two/three layers and record the performance gap on the validation set every hundred updates. We then plot those curves in this figure.} 
\label{Fig.what}
\vspace{-4mm}
\end{figure*}
To study this question, we 
fine-tune only the last one/two/three layers to provide strong stability for $H_{cross}^{pre}$. 
Figures~\ref{Fig.when} and~\ref{Fig.what} show that fine-tuning only the last few layers delays the first appearance of the performance gap and clearly decreases the performance gap. Also, the fewer layers are fine-tuned, the smaller the gap is. However, (1) a great gap still exists even if we only fine-tune the last layer (e.g., 9\% difference on the XNLI dataset). That means \textbf{avoiding the forgetting of the pre-trained $H_{cross}^{pre}$ can reduce the gap to some extent, but cannot solve the problem entirely.} (2) Fine-tuning fewer layers makes the overall performance drops significantly as the model does not have enough space to learn $H_{task}^{new}$ (see Table 1). That means a smaller performance gap is not equal to better overall performance and we need to consider the plasticity too.
\begin{table*}[h!]
\centering
\renewcommand{\arraystretch}{1.1}
\footnotesize
\scalebox{0.80}{
\renewcommand\tabcolsep{0.6mm}
\begin{tabular}{c|c|c|c|c||c|c|c||c|c|c}
\hline
{\bf Dataset}&{Baseline}&{Last one}&{Last two}&{Last three}&{Freeze four}&{Freeze attention}&{Freeze feed-forward}&{Enlarge two}&{Enlarge attention}&{Enlarge feed-forward}\\
\hline

XNLI &74.7{\scriptsize$\pm$0.2}&60.5{\scriptsize$\pm$0.2}&67.5{\scriptsize$\pm$0.3}&71.5{\scriptsize$\pm$0.4}&74.8{\scriptsize$\pm$0.1}&75.0{\scriptsize$\pm$0.1}&75.2{\scriptsize$\pm$0.3}&74.5{\scriptsize$\pm$0.2}&75.1{\scriptsize$\pm$0.4}&75.0{\scriptsize$\pm$0.3}
\\
\hline
MLQA&64.7{\scriptsize$\pm$0.3}&33.2{\scriptsize$\pm$0.7}&48.9{\scriptsize$\pm$0.3}&43.4{\scriptsize$\pm$0.1}&52.5{\scriptsize$\pm$0.5}&64.8{\scriptsize$\pm$0.2}&66.3{\scriptsize$\pm$0.3}&66.4{\scriptsize$\pm$0.3}&66.1{\scriptsize$\pm$0.5}&65.8{\scriptsize$\pm$0.6}
\\
\hline
NER&61.2{\scriptsize$\pm$0.1}&40.1{\scriptsize$\pm$0.2}&48.9{\scriptsize$\pm$0.3}&52.8{\scriptsize$\pm$0.5}&60.4{\scriptsize$\pm$0.6}&61.5{\scriptsize$\pm$0.3}&61.8{\scriptsize$\pm$0.1}&59.7{\scriptsize$\pm$0.1}&61.3{\scriptsize$\pm$0.2}&60.5{\scriptsize$\pm$0.2}
\\
\hline
\end{tabular}
}
\caption{Performance on XNLI, MLQA, and NER datasets in the zero-shot setting. All values are the averages of four different seeds. For the baseline, we directly fine-tune the pre-trained model. {\color{black}In the 'Last one/two/three' experiments, we only fine-tune the last one/two/three layers respectively. In the 'Freeze four'/'Freeze attention'/'Freeze feed-forward' experiments, we freeze the weights/attention weights/feed-forward weights in the first four layers. In the 'Enlarge two'/'Enlarge attention'/'Enlarge feed-forward' experiments, we enlarge the learning rate of the weights/attention weights/feed-forward weights in the last two layers by multiplying it with 10.} }
\vspace{-3mm}
\label{tab:gapattfeed}
\end{table*}
\section{Method}
This section proposes a method to avoid the forgetting of cross-lingual knowledge $H_{cross}^{pre}$ (stability) and to encourage the forgetting of task knowledge for the pre-training task $H_{task}^{pre}$ and learn new task's knowledge $H_{task}^{new}$ (plasticity). The core is to set different learning rate policies for the weights of the model based on both the layer's location and the training phase. 

\subsection{Reducing Forgetting of Cross-Lingual Knowledge with Fine-tuning slow} 
We consider the protection of $H_{cross}^{pre}$ first. The key challenge here is to strike a balance between maintaining  $H_{cross}^{pre}$ and learning $H_{task}^{new}$. Based on the above analysis, we propose a fine-tuning slow algorithm consisting of the following two training policies and apply them to different sets.

\textbf{Policy I: Avoiding drastic update of weights related to cross-lingual knowledge in the first fine-tuning phase $P_1$.} The performance gap increases quickly in $P_1$ and $H_{cross}^{pre}$ in weights are forgetting quickly. That is because the loss in this phase drops drastically and gives a big gradient update for each weight to update, and the stability for $H_{cross}^{pre}$ is not enough. So our goal here is to reduce the update of weights related to $H_{cross}^{pre}$ in this stage by multiplying their learning rate with a learning rate multiplier $K=c_1$ ($c_1<1$). 

\textbf{Policy II: Adjusting the learning rate of the key weights for cross-lingual transfer dynamically in the second fine-tuning phase $P_2$.} 
After the first phase, the gap increases slowly and our goal is to make the weights adapt to the downstream task and to avoid forgetting cross-lingual knowledge. So we set $K=1$ for weights related to $H_{cross}^{pre}$ to provide more plasticity and dynamically adjust the learning rate of the key weights related to $H_{cross}^{pre}$ additionally to avoid the forgetting. Our idea for the key weights is to provide more plasticity for them when the loss drops quickly to learn a new task and to provide more stability (avoiding unnecessary forgetting) when the loss drops slowly. 
We propose to set a dynamic multiplier $K$ based on the learning curve for the key weights of $H_{cross}^{pre}$. Assume that $\mathcal{L}_t$ is the training loss at the $t$th iteration and $\phi(\mathcal{L}_t)\in [0,1]$ is a function reflecting the tendency of the learning curve. The bigger the $\phi(\mathcal{L}_t)$ is, the faster the loss drops. Then we have $K=R(\phi(\mathcal{L}_t))$, where $R$ is a monotonic function. In this way, when the loss of the model drops quickly to adapt to the new task, $K$ is also bigger to provide more plasticity. 
Note that policy II in $P_2$ has no conflict with policy I as the drastic loss drop in $P_1$ is not desirable for the weights related to $H_{cross}^{pre}$ to adapt to the task.  

\textbf{Layers to apply policies I and II.} If re-initializing the weights in one layer obviously drops the performance across three datasets, we denote the weights in this layer belong to $S^{I}_\theta$. If freezing the weights in one layer improves the performance, we denote the weights in this layer belong to $S^{II}_\theta$ ($S^{II}_\theta \in S^{I}_\theta$). The latter is usually more important for cross-lingual transfer. Based on the re-initializing/freezing experiment (see Figure~\ref{Fig.where}),
we know that weights in the first 10 layers belong to $S^{I}_\theta$ and weights in the first 4 layers belong to $S^{II}_\theta$. 
For policy I, we apply it to $S^{I}_\theta$ as we do not want to forget cross-lingual knowledge $H_{cross}^{pre}$ due to the big gradient updates. 

\textbf{Attention vs Feed-forward}
To further investigate the best choice of protecting weights in the first four layers, we conduct experiments to freeze all weights/all weights of the multi-head layer/all weights of the feed-forward layer in the first four layers. The results in the second part of Table 1 show that freezing all weights of the feed-forward layer in the first four layers achieves the best performance over three datasets. With the additional re-initialization experiments (see Table~\ref{tab:re-four} in Appendix C), we find that is because the weights of the feed-forward layer are the most important weights for cross-lingual transfer in the first four layers. So we apply policy II to the weights of the feed-forward layer in $S^{II}_\theta$ as we want to protect $H_{cross}^{pre}$ and to provide more plasticity to learn $H_{new}^{task}$.



\subsection{Encouraging Forgetting of Pre-training Task Knowledge with Learning Fast} 
\label{sec.forget}
As shown earlier, the \textbf{pre-training task} knowledge $H_{pre}^{task}$ is usually useless or even harmful to the downstream task. 
Here we design an algorithm to utilize big gradient updates 
 (naturally happen in the first phase or are created by enlarging the learning rate) to encourage the model to forget $H_{pre}^{task}$ and to learn better 
 downstream task's knowledge $H_{new}^{task}$. 
 We refer to this as the fine-tuning fast algorithm consisting of two training policies and apply them to different sets:

\textbf{Policy III: Do not slow down the update of the weights related to $H_{pre}^{task}$ in the first fine-tuning phase $P_1$.} In $P_1$, {\color{black}the model is actively looking for a point that can reduce the loss drastically and has enough energy to break the limitation of the pre-trained knowledge $H_{pre}^{task}$.} So we allow the model to update the weights related to $H_{pre}^{task}$ in this phase without lessening their learning rate. 

\textbf{Policy IV: Increasing the learning rate of the key weights related to $H_{pre}^{task}$ in the second fine-tuning phase $P_2$.} In $P_2$, the loss drops gradually and the model finally needs to converge to local minima. But the model may not stop learning the new task's knowledge $H_{new}^{task}$. To verify this, we use the representation similarity metric CKA~\cite{pmlr-v97-kornblith19a} to measure the similarity of the representation of the current training data batch to the pre-trained model and to the current training model. Figure \ref{Fig.cka} in Appendix D shows that the similarity of the hidden representation from the last two layers is still dropping in the second phase (except the NER dataset) and the model is striving to learn a better task representation that is different from the pre-trained one. But the loss becomes small and drops slowly in $P_2$ and the model doesn't have enough energy \cite{pezeshki2021gradient} to forget the pre-training task knowledge and to learn $H_{new}^{task}$. So if the representation similarity of the last two layers is still dropping in $P_2$, we encourage the model to update the key weights relevant to the task knowledge by multiplying their learning rate with a learning rate multiplier $K=c_2$ ($c_2>1$). 

\textbf{Layers to apply policies III and IV.}
Based on the re-initializing experiment (Figure 2),
we know that re-initializing the weights in the last two layers improves the performance or drops the performance slightly. That means that the weights in the two layers have little cross-lingual knowledge and have $H_{pre}^{task}$ which has a negative effect on the learning of the downstream task. We denote the set of weights that has this property as $V^{I}_\theta$ and apply policy III to it. 

\textbf{Attention vs Feed-forward}
In the second phase, the model is trying to learn a better task representation and needs to converge to a stable point. So enlarging the learning rate of all weights in $V^{I}_\theta$ may not be the best choice (e.g., disturbing the convergence). 
To investigate the best choice of weights in the last two layers, we conduct experiments with an increased learning rate of different weight sets. 
Based on the results of the third part of Table 1, we find that increasing the learning rate of all weights in the attention layer of the last two layers achieves the best performance. That implies the weights of the attention layer are the key weight in the learning of the downstream task and that not changing the learning rate of other weights in the last two layers provides much stability. So we denote the weights of the attention layer in $V^{I}_\theta$ as $V^{II}_\theta$ and apply policy IV to it. 


\subsection{Fine-tuning slow and Fast Algorithm}

Formally, a typical multi-lingual pre-trained model $F$ comprises a stack of $L$ transformer layers with each layer containing an attention head layer $l_\theta^{a}$ followed by a feed-forward
network $l_\theta^{f}$. 
At the $t$-th training iteration of the fine-tuning process, the updating rule of the weight $\theta_t$ of model $F$ based on our fine-tuning slow and fast algorithm is:

\begin{equation}
\small
    \theta_t=\left\{\begin{array}{lll}
     \theta_{t-1}-K\cdot r\nabla\theta_{t-1}    & \textit{if}\quad t\in P_1 \land \theta\in S^{I}_\theta||&\\
           & t in P_2 \land \theta\in V^{II}_\theta \bigcup S^{II}_\theta&\\
     \theta_{t-1}- r\nabla\theta_{t-1}    & \textit{otherwise}
     
    \end{array} \right.
    \label{eq.weight}
\end{equation}

where $r$ is the learning rate and $\nabla\theta_{t-1}$ is the weight modification calculated by the back-propagation algorithm. $S^{I}_\theta$, $S^{II}_\theta$, $V^{I}_\theta$ and $V^{II}_\theta$ are the weight sets for the application of policy I, II, III, and IV respectively. We use $t\in P_1$ to identify if the $t$-th iteration belongs to the first phase $P_1$. The learning rate multiplier $K$ is determined by:
\begin{equation}
\small
    K=\left\{
    \begin{array}{lll}
        c_1 & \textit{if}\quad t\in P_1 \land \theta\in S^{I}_\theta &\\
        c_2 & \textit{if}\quad t\notin P_1 \land \theta\in V^{II}_\theta &\\
        R(\phi(\mathcal{L}_t)) 
         & \textit{if}\quad t\notin P_1 \land \theta\in S^{II}_\theta &
    \end{array}
    \right.
    \label{eq.k}
\end{equation}
where $c_1$ and $c_2$ are constant and $R(\phi(\mathcal{L}_t))$ is a monotonic function based on the function $\phi(\mathcal{L}_t)$ that can reflect the tendency of the learning curve. 
Our method maintains the stability for cross-lingual knowledge and increases the plasticity to adapt to the new task. We verify it in the following section.


\begin{table*}[h!]
\centering
\renewcommand{\arraystretch}{1.1}
\footnotesize
\scalebox{0.950}{

\renewcommand\tabcolsep{0.6mm}

\begin{tabular}{c|cccc|cccc|cccc}
\hline
{\bf Dataset}&{}&{}&{XNLI}&{}&{}&{}&{NER}&{}&{}&{}&{MLQA}&{}\\
\hline
{\bf M}&{0}&{5}&{10}&{20}&{0}&{5}&{10}&{20}&{0}&{5}&{10}&{20}\\
\hline
DF &74.7{\scriptsize$\pm$0.2}&75.1{\scriptsize$\pm$0.2}&75.4{\scriptsize$\pm$0.3}&75.5{\scriptsize$\pm$0.3}
&61.3{\scriptsize$\pm$0.2}&68.1{\scriptsize$\pm$0.1}&70.6{\scriptsize$\pm$0.1}&72.5{\scriptsize$\pm$0.1}
&64.7{\scriptsize$\pm$0.2}&64.8{\scriptsize$\pm$0.3}&64.8{\scriptsize$\pm$0.2}&64.9{\scriptsize$\pm$0.2}
\\
\hline
NosiyTune &74.9{\scriptsize$\pm$0.2}&75.1{\scriptsize$\pm$0.1}&75.5{\scriptsize$\pm$0.2}&75.6{\scriptsize$\pm$0.1}
&61.3{\scriptsize$\pm$0.1}&67.8{\scriptsize$\pm$0.3}&70.7{\scriptsize$\pm$0.2}&72.7{\scriptsize$\pm$0.2}
&64.8{\scriptsize$\pm$0.2}&64.8{\scriptsize$\pm$0.2}&64.9{\scriptsize$\pm$0.3}&65.0{\scriptsize$\pm$0.2}
\\
\hline
FS&75.2{\scriptsize$\pm$0.3}&75.5{\scriptsize$\pm$0.2}&75.5{\scriptsize$\pm$0.3}&76.0{\scriptsize$\pm$0.1}
&62.3{\scriptsize$\pm$0.2}&68.5{\scriptsize$\pm$0.2}&71.2{\scriptsize$\pm$0.3}&72.7{\scriptsize$\pm$0.2}
&66.1{\scriptsize$\pm$0.5}&66.3{\scriptsize$\pm$0.3}&66.4{\scriptsize$\pm$0.5}&66.8{\scriptsize$\pm$0.2}
\\
\hline
FF&75.0{\scriptsize$\pm$0.2}&75.4{\scriptsize$\pm$0.2}&75.6{\scriptsize$\pm$0.4}&75.9{\scriptsize$\pm$0.2}
&62.1{\scriptsize$\pm$0.2}&68.3{\scriptsize$\pm$0.3}&70.7{\scriptsize$\pm$0.1}&72.5{\scriptsize$\pm$0.2}
&66.4{\scriptsize$\pm$0.2}&66.5{\scriptsize$\pm$0.3}&66.5{\scriptsize$\pm$0.2}&66.7{\scriptsize$\pm$0.3}
\\
\hline
Our method &\bf 75.6{\scriptsize$\pm$0.1}&\bf75.7{\scriptsize$\pm$0.3}&\bf76.1{\scriptsize$\pm$0.2}&\bf76.5{\scriptsize$\pm$0.2}
&\bf62.5{\scriptsize$\pm$0.1}&\bf69.1{\scriptsize$\pm$0.2}&\bf71.7{\scriptsize$\pm$0.1}&\bf72.9{\scriptsize$\pm$0.1}
&\bf66.6{\scriptsize$\pm$0.2}&\bf66.8{\scriptsize$\pm$0.3}&\bf66.8{\scriptsize$\pm$0.2}&\bf67.0{\scriptsize$\pm$0.3}
\\
\hline
\end{tabular}

}
\caption{Performance on XNLI, MLQA, and NER datasets in the zero-shot and few-shot settings. All values are the averages of four different seeds. M is the number of few-shot training data for each non-source language. DF (directly fine-tuning the model), NoisyTune~\cite{wu2022noisytune}, FS (fine-tuning slow algorithm), and FF (fine-tuning fast algorithm) are the baselines.}
\vspace{-3mm}
\label{tab:accresults}
\end{table*}
\begin{table}[h!]
\centering
\renewcommand{\arraystretch}{1.1}
\footnotesize
\scalebox{0.950}{

\renewcommand\tabcolsep{0.6mm}

\begin{tabular}{c|cccc}
\hline
{\bf Language}&{en}&{fr}&{de}&{avg}\\
\hline
DF &70.2{\scriptsize$\pm$0.2}&69.1{\scriptsize$\pm$0.5}&70.0{\scriptsize$\pm$0.1}&69.7{\scriptsize$\pm$0.2}
\\
\hline
Our method &70.5{\scriptsize$\pm$0.2}&70.4{\scriptsize$\pm$0.1}&70.7{\scriptsize$\pm$0.2}&70.5{\scriptsize$\pm$0.2}
\\
\hline
\end{tabular}

}
\caption{Accuracy performance on Large QAM dataset in the zero-shot setting. All values are the averages of four different seeds.  'avg' is the average performance over all target languages.}
\vspace{-3mm}
\label{tab:QAM_results}
\end{table}
\begin{table}[h]
\centering
\renewcommand{\arraystretch}{1.1}
\footnotesize
\scalebox{0.80}{

\renewcommand\tabcolsep{0.6mm}

\begin{tabular}{c|cc|cc}

\hline
\multicolumn{1}{c|}{Method}&\multicolumn{2}{c|}{Baseline}&\multicolumn{2}{c}{Our method}\\
\hline
{\bf Performance}&{source}&{non-source}&{source}&{non-source}\\
\hline
 XNLI&84.8{\scriptsize$\pm$0.3}&74.0{\scriptsize$\pm$0.3}&85.6{\scriptsize$\pm$0.2} (+0.8)&74.9{\scriptsize$\pm$0.2} (+0.9)
\\
\hline
NER &82.3{\scriptsize$\pm$0.2}&60.7{\scriptsize$\pm$0.3}&82.3{\scriptsize$\pm$0.2} (+0.0)&62.0{\scriptsize$\pm$0.1} (+1.3)
\\
\hline
MLQA&79.4{\scriptsize$\pm$0.3}&62.2{\scriptsize$\pm$0.2}&80.5{\scriptsize$\pm$0.2} (+1.1)&64.3{\scriptsize$\pm$0.2} (+2.1)
\\
\hline

\end{tabular}

}
\caption{Source and non-source languages' performance in the zero-shot setting. For the baseline, we directly fine-tune the model. All values are the averages of four different seeds. }
\vspace{-3mm}
\label{tab:sourceresults}
\end{table}
\begin{table*}[h!]
\centering
\renewcommand{\arraystretch}{1.1}
\footnotesize
\scalebox{0.825}{

\renewcommand\tabcolsep{0.6mm}

\begin{tabular}{c|ccccc|ccccc|cccc}

\hline
{\bf Hyper-parameter}&{}&{}&{$c_1$}&{}&{}&{}&{}&{$c_2$}&{}&{}&{}&{}&{$r$}&{}\\
\hline
{\bf Value}&{0.5}&{0.1}&{0.01}&{0.001}&{0}&{5}&{10}&{15}&{20}&{100}&{1}&{2}&{3}&{4}\\
\hline
XNLI &75.3{\scriptsize$\pm$0.1}&75.1{\scriptsize$\pm$0.2}&75.6{\scriptsize$\pm$0.1}&75.5{\scriptsize$\pm$0.2}
&75.5{\scriptsize$\pm$0.1}&75.1{\scriptsize$\pm$0.2}&75.6{\scriptsize$\pm$0.1}&75.0{\scriptsize$\pm$0.1}&74.9{\scriptsize$\pm$0.4}&71.8{\scriptsize$\pm$0.3}&75.2{\scriptsize$\pm$0.1}&75.4{\scriptsize$\pm$0.1}
&75.6{\scriptsize$\pm$0.1}&75.4{\scriptsize$\pm$0.3}
\\
\hline
NER &62.1{\scriptsize$\pm$0.2}&62.2{\scriptsize$\pm$0.2}&62.5{\scriptsize$\pm$0.1}&62.3{\scriptsize$\pm$0.1}
&62.0{\scriptsize$\pm$0.2}& & &-&&
&62.1{\scriptsize$\pm$0.1}&62.3{\scriptsize$\pm$0.1}&62.5{\scriptsize$\pm$0.1}
&62.2{\scriptsize$\pm$0.2}
\\
\hline
MLQA&66.1{\scriptsize$\pm$0.}&66.5{\scriptsize$\pm$0.1}&66.6{\scriptsize$\pm$0.2}&66.3{\scriptsize$\pm$0.3}
&66.3{\scriptsize$\pm$0.3}&66.1{\scriptsize$\pm$0.2}&66.6{\scriptsize$\pm$0.2}&66.4{\scriptsize$\pm$0.1}
&66.0{\scriptsize$\pm$0.2}&27.3{\scriptsize$\pm$0.7}&66.4{\scriptsize$\pm$0.3}&66.4{\scriptsize$\pm$0.2}
&66.6{\scriptsize$\pm$0.2}&66.2{\scriptsize$\pm$0.2}
\\
\hline

\end{tabular}

}
\caption{Performance on XNLI, NER, and MLQA datasets in the zero-shot with different hyper-parameter values of $c_1$,$c_2$, and $r$. All values are the averages of four different seeds. We don't record the performance on the NER dataset with different $c_2$ here. The reason is: as a low-level task, the NER task is similar to the pre-training task and its task representation does not continue to be far from the pre-trained one in the second phase (see Figure~\ref{Fig.cka}) and so we don't apply policy IV on this dataset.}
\vspace{-3mm}
\label{tab:ablationresults}
\end{table*}
\section{Experiment}
We now use  three downstream tasks: Named Entity Recognition (NER), Question Answering (QA), and Natural Language Inference (NLI), to experimentally
evaluate the performance of our proposed \textbf{Fine-tuning slow and fast algorithm} under the zero-shot and few-shot settings. 

\subsection{Experiment Setup}
\textbf{Datasets}: We adopt the NER \cite{hu2020xtreme}, MLQA \cite{lewis2019mlqa}, and XNLI \cite{conneau2018xnli} datasets from the XTREME benchmark \cite{hu2020xtreme}
for NER, QA, and NLI respectively. The details of the datasets and training details are listed in Section 2. 

\textbf{Zero-shot and Few-shot settings.} We define the zero-shot setting as fine-tuning a pre-trained model for a downstream task
using its labeled data in one source language (e.g. English). Then we apply the fine-tuned model to all target languages. We define the few-shot setting as fine-tuning a pre-trained model for a downstream task
using its labeled data in one source language (e.g., English) and a few labeled data from other languages. All labeled data are mixed to form a training dataset and then we use it to fine-tune the pre-trained model. {\color{black}For the source of the few-shot data, we split the original validation set into the few-shot data group and the new validation set. Note that the number of data points in the validation set is usually larger than 5000. So extracting the few-shot data from the validation set does not influence its validation function.} 

\textbf{Baselines}: 
(1) Directly Fine-tuning (DF) the model with the English training corpus; (2) NoisyTune~\cite{wu2022noisytune}, which prevents LMs
from overfitting the data in pre-training and reducing the gap between pre-training
and downstream tasks by adding a small
amount of noise to perturb the LM parameters before
fine-tuning. (3) Fine-tuning slow algorithm (FS), which fine-tunes the model with the fine-tuning slow algorithm. (4) Learning Fast algorithm (FF), which fine-tunes the model with the fine-tuning fast algorithm. 

\textbf{Choice of adaptive multiplier $R(\phi(\mathcal{L}_t))$ and $t\in P_1$, and hyper-parameters:}
For $R(\phi(\mathcal{L}_t))$ in Eq.~\ref{eq.k}, we first calculate the average value of the losses in the recent 100 iterations as $\frac{\mathcal{L}_{t-100:t}}{100}$ and the losses in the 100 iterations prior to the recent 100 iterations  as $\frac{\mathcal{L}_{t-200:t-100}}{100}$. Then we define $\phi(\mathcal{L}_t)=\frac{\mathcal{L}_{t-100:t}}{\mathcal{L}_{t-200:t-100}}$. When the loss drops quickly ($\mathcal{L}_{t-200:t-100}\gg\mathcal{L}_{t-100:t}$), $\phi(\mathcal{L}_t)$ is close to 0. And when the loss drops slowly, $\phi(\mathcal{L}_t)$ is close to 1. We do not use $\frac{\mathcal{L}_{t}}{\mathcal{L}_{t-1}}$ to represent $\phi(\mathcal{L}_t)$ as the losses in adjacent iterations usually do not have a big difference and so it cannot accurately describe the tendency. 
Then $R(\phi(\mathcal{L}_t))$ is defined as:
\begin{equation}
R(\phi(\mathcal{L}_t))=\max(1-\phi(\mathcal{L}_t)^{r},0)
    \label{eq.R}
\end{equation}
where $r$ is a hyper-parameter. When the loss drops quickly, $R(\phi(\mathcal{L}_t))$ is close to 1 and gives the parameters more plasticity to adapt to the new task and vice versa. We choose $\frac{\mathcal{L}_{t-200:t-100}-\mathcal{L}_{t-100:t}}{100}>0.1$ to represent $t\in P_1$ in Eq.~\ref{eq.weight} as the loss drops very quickly in this case and the model needs policies I and III to protect $H^{pre}_{cross}$ and learn $H^{new}_{task}$. We set $r$ as 3. We set $c_1$ and $c_2$ (Eq.~\ref{eq.k}) as 0.01 and 10 respectively. The ablation study is in Section 4.6.

\subsection{Results of Zero-Shot Fine-Tuning}
To evaluate the zero-shot performance of our method and baselines, we record the average F1 score performance (mean and standard deviation) of all target languages for the MLQA and NER datasets and the average accuracy for the XNLI dataset. The results are reported in Table 2, which shows 
that our method achieves highly superior results to the baseline methods.

\subsection{Results of Few-Shot Fine-Tuning}
The results of few-shot fine-tuning performance are reported in Table~\ref{tab:accresults}, which shows that:
(1) with the increasing number of few-shot data per language, the performance improves as the model can learn better cross-lingual task representation from the multi-lingual training corpus. (2) Our method still outperforms baselines obviously as it protects the pre-trained cross-lingual knowledge $H_{cross}^{pre}$ and forgets the pre-training task's knowledge $H_{task}^{pre}$. 

\subsection{Large Training Corpus Fine-Tuning}
We collect and construct a QAM task dataset \cite{liang2020xglue} with 12 million English training data points by a business search engine. QAM classification task aims
to predict whether a <question, passage> pair is a QA pair.   {\color{black}Zero-shot fine-tuning model on a corpus like this is more challenging as the model is easier to forget $H_{cross}^{pre}$.} From Table~\ref{tab:QAM_results}, we observe that our method outperforms the baseline obviously, which shows that our method also works well with a large dataset. {\color{black}From Figure~\ref{fig.qam} in Appendix E, we find that (1) the performance of non-source languages firstly increases and then drops (due to forgetting) (2) the performance of source language increases during the whole training process and the gap becomes larger in the later phase. (3) our method reduces the gap by protecting cross-lingual knowledge and improves the performance of non-source languages. More analyses and training details are in Appendix E.}

\subsection{Analysis of the Influence on Source Language and Non-Source Language}
Our method improves both source and non-source languages' performance in Table~\ref{tab:sourceresults}. Additionally, it helps non-source languages more, as it protects the cross-lingual knowledge in $H^{pre}_{cross}$, leading to a significant improvement in non-source languages' performance. We report each language's performance on the XNLI dataset in Appendix F.

\subsection{Analysis of the Influence of Learning Rate Multiplier Hyper-Parameters $c_1$, $c_2$ and $r$}
We ablate on the learning rate multiplier's hyper-parameters $c_1$, $c_2$ (Eq.~\ref{eq.k}) and $r$(Eq.~\ref{eq.R}). We analyze their influence by setting different values for them and recording the final overall performance. From Table~\ref{tab:ablationresults}, we observe that reducing the value of $c_1$ from 0.5 to 0 improves performance initially (by protecting $H^{pre}_{cross}$) but then decreases performance (due to lack of enough plasticity). We set $c_1$ as 0.01 as it strikes a balance between stability and plasticity. Increasing the value of $c_2$ from 5 to 100 improves performance initially (by learning better $H^{new}_{task}$) but then decreases performance (due to lack of stability to converge). So we set $c_2$ to 10.  Increasing $r$ from 1 to 4 improves performance initially (as the adaptive multiplier is closer to 1 and provides more stability) but then decreases performance (due to lack of enough plasticity). So we set $r$ as 3. Our choice is consistent across the three datasets, showing our method's robustness.

\subsection{Analysis of the Influence of the Usage Order of Weight Sets $S^{I}_\theta$,
$S^{II}_\theta$,$V^{I}_\theta$ and $V^{II}_\theta$.}
{\color{black}To further verify the effectiveness of our method in applying different weight sets for each policy, we conduct experiments that apply both Policies I and II to only $S^{I}_\theta$/$S^{II}_\theta$ respectively, as well as experiments that apply Policy I to $S^{II}_\theta$ and Policy II to $S^{I}_\theta$. We also perform similar experiments for sets $V^{I}_\theta$ and $V^{II}_\theta$. From Table~\ref{tab:order} in Appendix G, we find that our method achieves the best performance by making a good balance between plasticity and stability. Further analyses are in Appendix G.}

\section{Related Work}
\label{sec.related}
{\color{black}
\textbf{Fine-tuning multilingual language models (MLLMs).} Recent MLLM systems (e.g., mBERT \citep{devlin2018bert}, mT5\citep{xue2020mt5}, and XLM-R \cite{conneau2019unsupervised}) have shown strong zero-shot transfer ability to non-source languages when fine-tuning with only a source language corpus. However, the performance gap between the source language and non-source languages is still large. Most previous works focus on learning robust task representation\cite{fang2021filter,zheng2021consistency,jiang2022rose} and strong pre-trained cross-lingual representation \cite{chi2020infoxlm,wang2020negative,ouyang2020ernie,hu2020explicit}. But they haven't analyzed the performance gap in fine-tuning and the relation between forgetting and the gap. We fill this research gap. 

\textbf{Forgetting in continual learning.}
 Continual learning aims to design algorithms to learn tasks incrementally. Its main challenge is forgetting. Many methods have been proposed to reduce forgetting. (1) Regularization methods\cite{kirkpatrick2017overcoming,chen2020recall,li2022overcoming,lee2021sequential} penalize the changes on important weights for previous tasks, (2) Replay methods\cite{buzzega2020dark,rolnick2019experience,wang2022memory} re-train a few previous samples with new task's data, and (3) Parameter-fixed methods\cite{vidoni2020orthogonal,he2021effectiveness,xu2021raise} protect parameters learned for previous tasks. The regularization method needs to estimate the important weight during pre-training and the replay method needs to store data from the pre-training corpus. But we usually don't have those during the fine-tuning phase. Moreover, all of those methods focus on avoiding forgetting but we find that pre-training task knowledge is not suitable for the adaptation of downstream tasks and those methods cannot get rid of it. We propose a novel method that controls the forgetting effect to avoid the forgetting of cross-lingual knowledge and encourage the forgetting of pre-training task knowledge to learn better new task knowledge. Our method is orthogonal to previous works in cross-lingual fine-tuning.} 

\section{Conclusion}
{\color{black} This paper first analyzed when the performance gap arises and where the important cross-lingual knowledge is and reduced the lower bound of the performance gap by avoiding forgetting cross-lingual knowledge. Based on our analysis, a novel method is proposed to control the forgetting effect in fine-tuning a multi-lingual pre-trained model. We verify its effectiveness over multiple datasets and settings.}

\section{Limitations}
{\color{black} Although we believe that controlling the forgetting in the fine-tuning phase to avoid  forgetting cross-lingual/general knowledge and to reduce the negative interference from misaligned pre-training tasks and downstream tasks can benefit other fine-tuning settings (e.g. Multi-task setting), we have not yet investigated these settings. In the future, we will try to propose a more general method for fine-tuning a large pre-trained model across various settings.}
\appendix

\bibliography{anthology,custom}
\bibliographystyle{acl_natbib}

\appendix
\section{ Performance gap of the MLQA dataset and NER dataset}
\begin{figure*}[h]
\centering 
\vspace{-2mm}

\includegraphics[height=0.4\textwidth,width=0.75\textwidth]{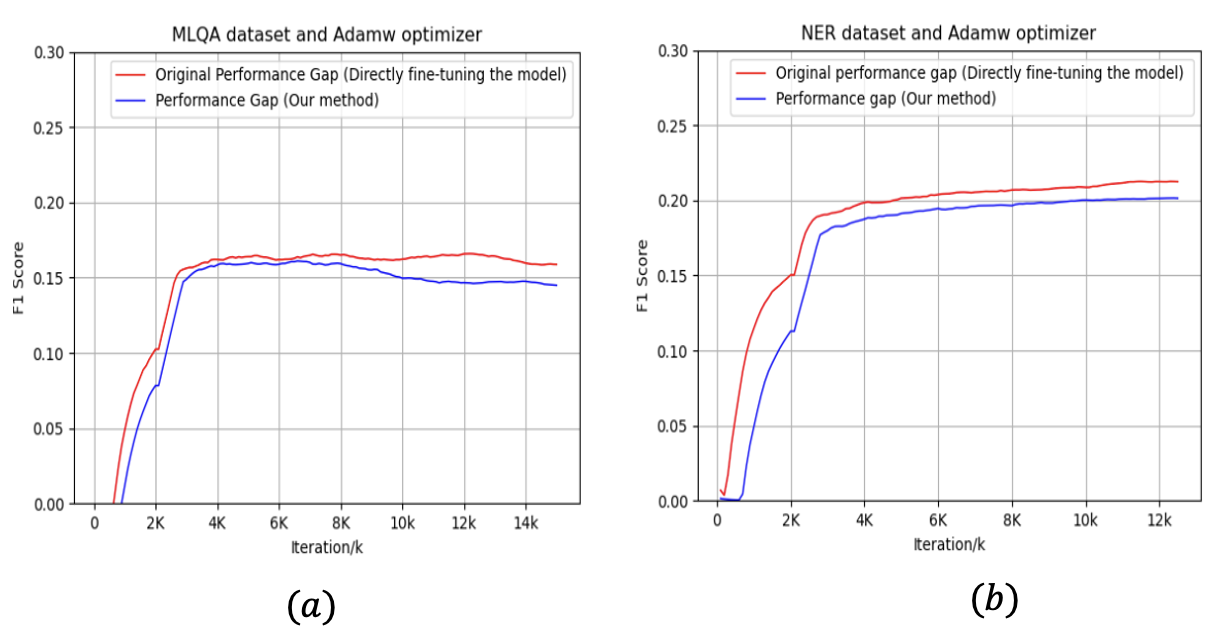} 
\vspace{-3mm}
\caption{The performance gap $P_{\hat{s}}-P_{S/\hat{s}}$  every hundred updates on the MLQA and NER datasets. 
'Original performance gap' means that we directly fine-tune the model, and 'our method' means that we use the Fine-tuning slow and fast algorithm to fine-tune the model.}
\label{Fig.gap_2}
\vspace{-4mm}
\end{figure*}
From Figure~\ref{Fig.gap_2}, we observe that (1) our method delays the first appearance  of the performance gap. (2) our method has a lower performance gap compared to the baseline.
\section{Details of the datasets and model.}
\textbf{XNLI} is a cross-lingual textual entailment dataset.
In this dataset, we use the MultiNLI \cite{williams2017broad} training data (English) to fine-tune the pre-trained model and then test the fine-tuned model with all 15 languages. 

\textbf{MLQA} is a multilingual machine reading comprehension task for question answering. The test performance gap is the gap between the F1 score of the source (English) and the average F1 score of the other six target languages (\textit{Arabic, German, Spanish, Hindi, Vietnamese} and \textit{Chinese}). 

\textbf{NER} is a named entity recognition task and we use the Wikiann~\cite{pan2017cross} dataset. The metric is the F1 score. We use the balanced train, dev, and test splits in \cite{rahimi2019massively}.

Following~\cite{hu2020xtreme}, the fine-tuning batch size is 32.
 We use the Adam optimizer with warm-up and learning rate
5e-6.
For XNLI, we fine-tune the model with the English corpus for 10 epochs and evaluate it on the English dev
set every 3k steps to select the best model.
For NER, we fine-tune 20 epochs. For MLQA, we follow BERT \cite{devlin2018bert} for SQuAD \cite{rajpurkar2016squad} and set
the learning rate to 3e-5, batch size to 12, and we train the model for
2 epochs. 

We select the model with the best
of the average result on the dev sets of all languages. 

XLM-R has 550 million parameters and we run the experiments with GPU A100.
\section{Re-initialization experiments for the weights in the first four layers}
\begin{table*}[h!]
\centering
\renewcommand{\arraystretch}{1.1}
\footnotesize
\scalebox{0.80}{

\renewcommand\tabcolsep{0.6mm}
\begin{tabular}{c|c|c|c|c}

\hline
{\bf Dataset}&{Baseline}&{Re-initialize four}&{Re-initialize attention}&{Re-initialize feed-forward}\\
\hline

XNLI &74.7{\scriptsize$\pm$0.2}&64.0{\scriptsize$\pm$0.1}&68.1{\scriptsize$\pm$0.1}&67.6{\scriptsize$\pm$0.2}
\\
\hline
MLQA&64.7{\scriptsize$\pm$0.3}&28.7{\scriptsize$\pm$0.3}&50.1{\scriptsize$\pm$0.1}&46.0{\scriptsize$\pm$0.1}
\\
\hline
NER&61.2{\scriptsize$\pm$0.1}&45.1{\scriptsize$\pm$0.2}&52.5{\scriptsize$\pm$0.2}&45.5{\scriptsize$\pm$0.2}
\\
\hline

\end{tabular}

}
\caption{Performance on XNLI, MLQA, and NER datasets in the zero-shot setting. All values are the averages of four different seeds. For the baseline, we directly fine-tune the pre-trained model. {\color{black} In the 'Re-initialize four', 'Re-initialize attention' and 'Re-initialize feed-forward' experiments, we re-initialize the weights/attention weights/feed-forward weights in the first four layers and then we fine-tune the model. } }
\vspace{-3mm}
\label{tab:re-four}
\end{table*}
From Table~\ref{tab:re-four}, we find that (1) re-initializing weights in the first four layers reduces the overall performance as the cross-lingual knowledge is lost. (2) Re-initializing the feed-forward weights in the first four layers has a worse effect on the overall performance than re-initializing the attention weights in the first four layers. That means the feed-forward weights in the first four layers are more important than the attention weights in the first four layers for cross-lingual transfer.

\section{Figure for the CKA representation similarity}
\begin{figure*}[h]
\centering 

\includegraphics[height=0.3\textwidth,width=1\textwidth]{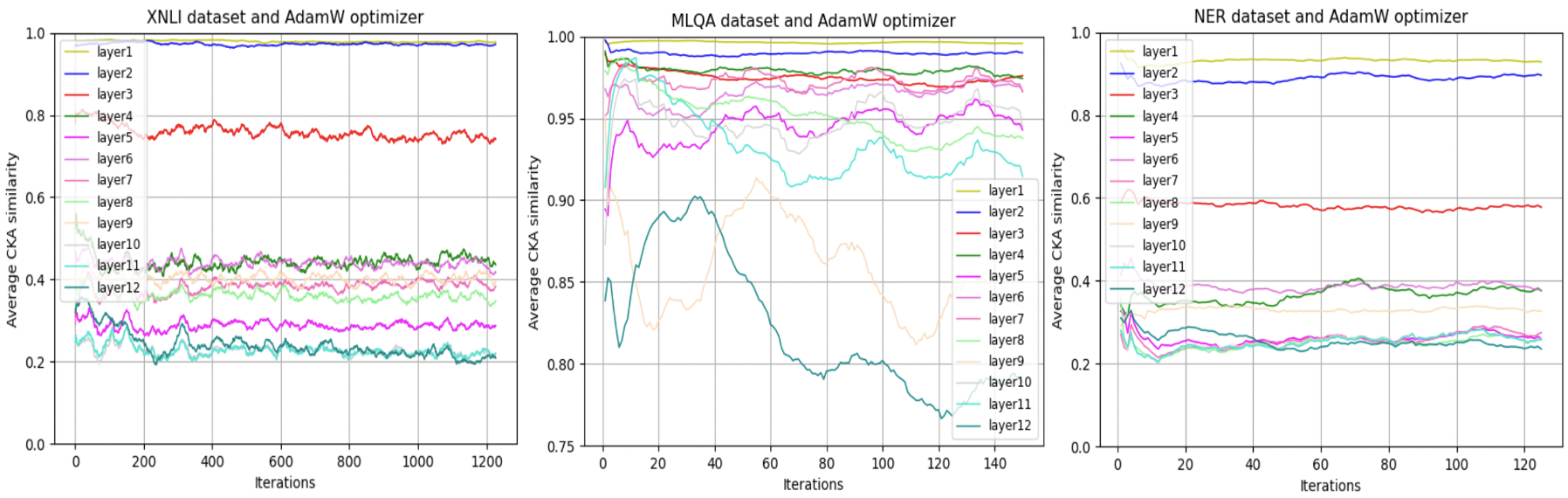} 
\vspace{-3mm}
\caption{For the pre-trained model and the current working model, we first calculate each layer's hidden representation on the current data batch, and then calculate and record the representation similarity of the hidden representations from the same layer in the pre-trained model and the working model every hundred updates over three different datasets. We plot the curves in this figure. A lower value (similarity) indicates a larger distance.}
\label{Fig.cka}
\vspace{-4mm}
\end{figure*}
\label{sec:appendix}
Figure~\ref{Fig.cka} describes the similarity of the hidden representations from the pre-trained model and the working model. We observe that the bottom layers usually have a higher similarity than the top layers, indicating that the top layers need larger adjustments to adapt to the downstream task. Also, we find that the similarity of the last two layers continues to decrease in the second phase over the MLQA and XNLI datasets, indicating that the model is still trying to learn a better task representation by modifying the weights in the last two layers during the second phase. 
\section{The performance analysis of the QAM dataset}

\begin{figure*}[h]
\centering 

\includegraphics[height=0.5\textwidth,width=1\textwidth]{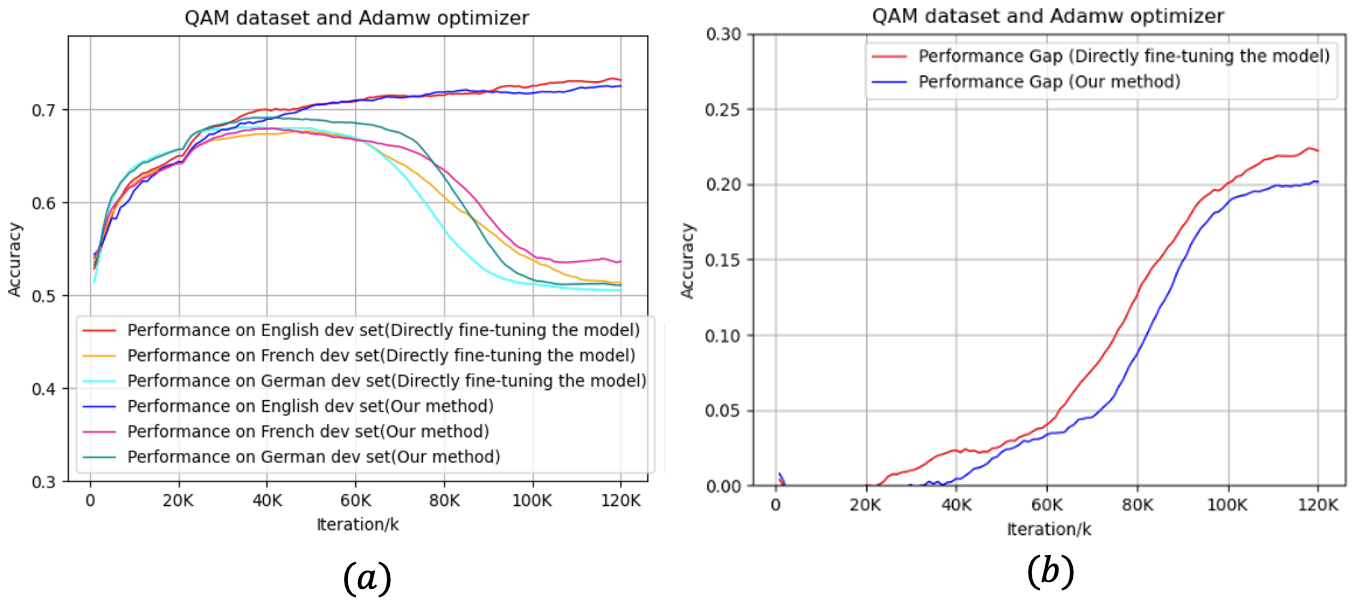} 
\vspace{-5mm}
\caption{We directly fine-tune the model on the QAM dataset as the baseline. English is the source language. German and French are the non-source language. We calculate and record their performance on the validation set every thousand updates.}
\label{fig.qam}
\vspace{-4mm}
\end{figure*}
We use the Adam optimizer 
with warm-up and learning rate 5e-6 to fine-tune the model with the English corpus for 1 epoch as the training corpus is big enough for the model to achieve the best performance by running one epoch. The batch size is 32.  We select the model with the best of
the average result on the dev sets of all languages (every 3k updates). 

We record the performance gap and each language's performance on the dev set every thousand updates and report the curve in the first 120k iterations (Figure~\ref{fig.qam}) as the best overall performance model is selected in the first 120k iterations. In the later iterations, the gap rises and the overall performance drops. As shown in Figure~\ref{fig.qam}, the tendency of the gap curve is monotonically increasing during the whole training process, and the main reason for this is the decline in the performance of the non-source languages. Our method improves overall performance by reducing the forgetting of cross-lingual knowledge.
\section{Analysis of the influence of our method for each language}
\begin{table*}[ht]
\centering
\renewcommand{\arraystretch}{1.1}
\footnotesize
\scalebox{0.80}{

\renewcommand\tabcolsep{0.6mm}

\begin{tabular}{c|ccc ccc ccc ccc ccc}

\hline
{\bf Language}&{ar}&{bg}&{de}&{el}&{en}&{es}&{fr}&{hi}&{ru}&{sw}&{th}&{tr}&{ur}&{vi}&{zh}\\
\hline
DF &72.2{\scriptsize$\pm$0.3}&78.0{\scriptsize$\pm$0.3}&77.3{\scriptsize$\pm$0.2}&76.0{\scriptsize$\pm$0.1}
&84.8{\scriptsize$\pm$0.3}&79.5{\scriptsize$\pm$0.2}&78.7{\scriptsize$\pm$0.3}&70.7{\scriptsize$\pm$0.2}
&76.1{\scriptsize$\pm$0.5}&65.1{\scriptsize$\pm$0.4}&72.7{\scriptsize$\pm$0.2}
&73.1{\scriptsize$\pm$0.1}&66.5{\scriptsize$\pm$0.5}&75.1{\scriptsize$\pm$0.1}&74.5{\scriptsize$\pm$0.4}
\\
\hline
NosiyTune &72.2{\scriptsize$\pm$0.3}&78.0{\scriptsize$\pm$0.3}&77.3{\scriptsize$\pm$0.2}&75.6{\scriptsize$\pm$0.1}
&84.8{\scriptsize$\pm$0.3}&79.5{\scriptsize$\pm$0.2}&78.7{\scriptsize$\pm$0.3}&70.7{\scriptsize$\pm$0.2}
&76.1{\scriptsize$\pm$0.5}&65.1{\scriptsize$\pm$0.4}&72.7{\scriptsize$\pm$0.2}
&73.1{\scriptsize$\pm$0.1}&66.5{\scriptsize$\pm$0.5}&75.1{\scriptsize$\pm$0.1}&74.5{\scriptsize$\pm$0.4}
\\
\hline
FS&73.0{\scriptsize$\pm$0.2}&78.4{\scriptsize$\pm$0.5}&77.6{\scriptsize$\pm$0.2}&76.6{\scriptsize$\pm$0.3}
&85.2{\scriptsize$\pm$0.2}&80.0{\scriptsize$\pm$0.1}&79.3{\scriptsize$\pm$0.1}&71.1{\scriptsize$\pm$0.1}
&76.8{\scriptsize$\pm$0.4}&65.4{\scriptsize$\pm$0.5}&72.9{\scriptsize$\pm$0.1}&74.0{\scriptsize$\pm$0.3}&66.5{\scriptsize$\pm$0.3}&75.7{\scriptsize$\pm$0.1}
&75.0{\scriptsize$\pm$0.3}
\\
\hline
FF&72.6{\scriptsize$\pm$0.4}&78.0{\scriptsize$\pm$0.2}&77.4{\scriptsize$\pm$0.1}
&76.1{\scriptsize$\pm$0.2}&85.0{\scriptsize$\pm$0.0}&79.1{\scriptsize$\pm$0.4}&78.5{\scriptsize$\pm$0.4}
&70.7{\scriptsize$\pm$0.2}&76.1{\scriptsize$\pm$0.4}&65.4{\scriptsize$\pm$0.4}&73.6{\scriptsize$\pm$0.1}
&73.8{\scriptsize$\pm$0.4}&67.2{\scriptsize$\pm$0.6}&75.4{\scriptsize$\pm$0.3}&74.9{\scriptsize$\pm$0.1}
\\
\hline
Our method &73.6{\scriptsize$\pm$0.2}&79.0{\scriptsize$\pm$0.2}&78.1{\scriptsize$\pm$0.3}&76.6{\scriptsize$\pm$0.2}
&85.6{\scriptsize$\pm$0.2}&80.2{\scriptsize$\pm$0.5}&79.5{\scriptsize$\pm$0.1}&71.5{\scriptsize$\pm$0.2}
&77.1{\scriptsize$\pm$0.3}&65.9{\scriptsize$\pm$0.3}&73.3{\scriptsize$\pm$0.1}
&74.2{\scriptsize$\pm$0.3}&67.1{\scriptsize$\pm$0.1}&76.2{\scriptsize$\pm$0.1}&75.9{\scriptsize$\pm$0.2}
\\
\hline
\end{tabular}
}
\caption{Each target language's performance on the XNLI dataset in the zero-shot setting. All values are the averages of four different seeds. }
\vspace{-3mm}
\label{tab:lanresults}
\end{table*}
From Table~\ref{tab:lanresults}, we find that the fine-tuning slow algorithm improves almost all languages' performance as it protects the cross-lingual knowledge.
The fine-tuning fast algorithm (learning the new task knowledge $H^{new}_{task}$) also improves the performance of the source language and some non-source languages as it provides a better task representation. Our method achieves the best performance over all languages, especially for some low-resource languages (e.g., sw and ur).
\section{Analysis of the influence of $S^{I}_\theta$,$S^{II}_\theta$,$V^{I}_\theta$, and $V^{II}_\theta$}
From Table~\ref{tab:order}, we observe that the 'Only $S^{I}_\theta$/$S^{II}_\theta$' experiments achieve worse performance than our method as they lack enough plasticity/stability for the cross-lingual knowledge, respectively. The '$S^{II}_\theta$ to $S^{I}_\theta$' experiment achieves the worst performance as it lacks enough stability in the first phase and lacks enough plasticity in the second phase. The 'Only $V^{I}_\theta$/$V^{II}_\theta$' experiments do not provide enough space for learning new tasks in the first phase/stability to converge in the second phase, respectively, so their performance is worse. The '$V^{II}_\theta$ to $V^{I}_\theta$' experiment has both disadvantages of the 'Only $V^{I}_\theta$/$V^{II}_\theta$' experiments, so its performance is also poor. We did not apply policy IV on the NER dataset, so we did not record the result of the '$V^{II}_\theta$ to $V^{I}_\theta$' experiment on the NER dataset.
\begin{table*}[h!]
\centering
\renewcommand{\arraystretch}{1.1}
\footnotesize
\scalebox{0.80}{

\renewcommand\tabcolsep{0.6mm}
\begin{tabular}{c|c|c|c|c||c|c|c}

\hline
{\bf Dataset}&{Our method}&{Only $S^{I}_\theta$}&{Only $S^{II}_\theta$}&{$S^{II}_\theta$to$S^{I}_\theta$}&{Only $V^{I}_\theta$}&{Only $V^{II}_\theta$}&{$V^{II}_\theta$to$V^{I}_\theta$}\\
\hline

XNLI &75.6{\scriptsize$\pm$0.1}&75.2{\scriptsize$\pm$0.1}&74.5{\scriptsize$\pm$0.1}&74.2{\scriptsize$\pm$0.2}&74.6{\scriptsize$\pm$0.3}&74.7{\scriptsize$\pm$0.2}&74.5{\scriptsize$\pm$0.2}
\\
\hline
MLQA&66.6{\scriptsize$\pm$0.2}&66.2{\scriptsize$\pm$0.2}&65.9{\scriptsize$\pm$0.3}&65.1{\scriptsize$\pm$0.1}&66.3{\scriptsize$\pm$0.3}&66.2{\scriptsize$\pm$0.2}&66.3{\scriptsize$\pm$0.1}
\\
\hline
NER&62.5{\scriptsize$\pm$0.1}&62.4{\scriptsize$\pm$0.3}&62.3{\scriptsize$\pm$0.2}&62.3{\scriptsize$\pm$0.2}&62.5{\scriptsize$\pm$0.1}&62.3{\scriptsize$\pm$0.1}&-
\\
\hline

\end{tabular}

}
\caption{Performance on XNLI, MLQA, and NER datasets in the zero-shot setting. All values are the averages of four different seeds. {\color{black}In the 'Only $S^{I}_\theta$/$S^{I}_\theta$' experiments, we apply both policies I and II to only $S^{I}_\theta$/ the weights of the feed-forward layer in $S^{II}_\theta$ respectively. In the '$S^{II}_\theta$to$S^{I}_\theta$' experiment, we apply policy I to the weights of the feed-forward layer in $S^{II}_\theta$ and apply policy II to $S^{I}_\theta$. In the 'Only $V^{I}_\theta$/$V^{II}_\theta$' experiments, we apply both policy III and IV to only $V^{I}_\theta$/$V^{II}_\theta$ respectively.
 In the '$V^{II}_\theta$to$V^{I}_\theta$' experiment, we apply policy III to $V^{II}_\theta$ and apply policy IV to $V^{I}_\theta$.} }
\vspace{-3mm}
\label{tab:order}
\end{table*}

\end{document}